\pdfoutput=1
\documentclass[11pt,onecolumn]{article}
\usepackage{cvpr}
\usepackage{amsmath, amssymb, graphicx, xfrac, amsfonts, amsbsy, algorithm, color, subfig, cite, url, multirow, hyperref}
\usepackage[noend]{algpseudocode}
\usepackage[section]{placeins}
\usepackage{float}

\cvprfinalcopy

\def\A {\mathbf{A}}

\def\Gd {\mathbf{G}}
\def\xb {\mathbf{x}}
\def\Xb {\mathbf{X}}
\def\Yb {\mathbf{Y}}

\def\x' {\xb'}
\def\xest {\widehat{\xb}}
\def\xbd{ \Yb^l_{\text{lidar}} }
\def\xbi{ \Yb^l_{\text{cam}} }
\def\y {\mathbf{y}}
\def\dO { \Omega' }
\def\tb {\pmb{\theta}}
\def\test {\widehat{\tb}}
\def\s {\mathbf{s}}
\def\sO {\s_{\Omega}}
\def\zO {\mathbf{z}_{\Omega}}

\def\G { \triangledown}
\def\g { \mathbf{D} }

\def\I{ \mathbf{ I } }
\def\IO{ \I_{\Omega} }
\def\O'{ \Omega'}

\def \SO3{ \mathbf{SO}(3) } 
\def \SE3{ \mathbf{SE}(3) } 

\def\R{\mathbb{R}}

\def\Acal{\mathcal{A}}

\def\argmin{\mathop{\mathrm{arg\,min}}}

\makeatletter
\def\BState{\State\hskip-\ALG@thistlm}
\makeatother

\begin{document}


\title{Motion Guided LiDAR-camera Self-calibration and Accelerated Depth Upsampling for Autonomous Vehicles}


\author{Juan Castorena, Gintaras V. Puskorius and Gaurav Pandey
	\thanks{J. Castorena, G. Puskorius and G. Pandey are at the Robotics and AI laboratory, Ford Motor Company, Dearborn, MI.
	}
}

\maketitle

\begin{abstract}
{\normalfont
This work proposes a novel motion guided method for targetless self-calibration of a LiDAR and camera and use the re-projection of LiDAR points onto the image reference frame for real-time depth upsampling. The calibration parameters are estimated by optimizing an objective function that penalizes distances between 2D and re-projected 3D motion vectors obtained from time-synchronized image and point cloud sequences. For upsampling, a simple, yet effective and time efficient formulation that minimizes depth gradients subject to an equality constraint involving the LiDAR measurements is proposed. Validation is performed on recorded real data from urban environments and demonstrations that our two methods are effective and suitable to mobile robotics and autonomous vehicle applications imposing real-time requirements is shown.
}
\end{abstract}




\section{Introduction}
\label{Sec:Introduction} 


A broad range of mobile robotic (e.g., autonomous vehicle) applications exploit fusion of information from sources of multiple modalities to reduce the uncertainties or limitations inherited when any of these sources are used individually. One of the most popular and effective complementary pair of sensors for robotic perception are cameras capturing RGB or IR information and depth systems (e.g., LiDAR) providing direct 3D information of the environment. Other systems that are heavy on computation but cheap in sensing have also been devised to obtain 3D information. For example some works \cite{Saxena05,Zbontar16, Kendall17} rely on learned models for depth prediction using deep networks and mono, stereo or multiple passive illumination camera sensors.
However, the performance of these significantly degrade in environments with poor illumination and extremes of structural variability (e.g., tunnels or hallways, vegetation). To overcome these passive sensing based limitations, methods that fuse camera and LiDAR information for depth estimation have been proposed \cite{Diebel05, He13, Lu15, Ferstl13}. These in general exploit the information gathered through the sensor measurements and uses priors or learned geometrical constraints that dictate how to fuse the information. 
%
For example, the early work proposed by Diebel and Thrun \cite{Diebel05} demonstrated the idea of constraining a Markov random field (MRF) based optimization to yield depth-maps with edges that co-occur with those from the camera image, assuming modalities have been co-registered. He et al. \cite{He13} proposed an image guided filtering approach to guide and improve edge selection in the reconstruction of upsampled depth. The work of Lu and Forsyth \cite{Lu15} proposed a segmentation based approach where sparse LiDAR points are used to segment  the image followed by smoothing of the sparse points within each of these segments to generate depth.  The sparse-promoting regularization based work of Ferstl et al. \cite{Ferstl13} proposed to use a weighted total generalized variation (TGV) formulation to promote consistency with edges from a high-resolution image in the depth upsampled reconstruction. More recent works have also exploited temporal information to further enhance reconstruction. 
For example, the work of Kamilov and Boufounos\cite{Kamilov17} formulated an inverse problem that uses a low-rank penalty on groups of 
related depth patches found from corresponding image patch similarity. The work of Degraux et al. \cite{Degraux17} instead learns group-sparse representations and uses total variation (TV) regularization for depth upsampled reconstruction. 

Methods that rely on fusing (in the optimal sense) information between multiple sensing modalities tend to be in general more robust than its individual sensor usage counterparts. Fusing, for example, LiDAR and camera measurements for depth upsampling predictions makes the reconstruction more invariant to illumination changes and extremes of environmental structure detail. However, most of these assume perfect alignment between data from the multiple modalities, an event that rarely occurs in practice. In fact, most fused based depth reconstruction methods are very sensitive to even small extrinsic miss-calibrations between the LiDAR-camera system. To this end, one of the first effective works to correct small miss-calibrations automatically and in an online fashion computes edges simultaneously in both modalities and uses a cost function of correlation to measure alignment between edges as in Levinson and Thrun \cite{Levinson.Thrun2013}. The work of Pandey et al. \cite{Pandey14} uses instead a mutual information function to measure the similarity between the reflectivity measurements from a LiDAR and the intensities from camera images assuming availability of reliable reflectivity measurements from a LiDAR. Other more recent automatic methods like Castorena et al. \cite{Castorena.Kamilov.Boufounos2016} considers a joint alternate optimization between data alignment and depth-upsampling reconstruction based on the idea that the better the alignment is the more accurate the depth reconstruction and vice-versa. Another method by Scott et al. \cite{Scott16} exploiting temporal information automatically attends specific time and place instances and formulates an objective function optimized to find the rigid body parameters that align the attended data. 

This paper is focused on both the problem of depth up-sampling reconstructions for robotic applications imposing real-time requirements and on the problem of extrinsic self-calibration of a camera-LiDAR system. The contributions of the work, detailed in Section \ref{Sec:proposedApproach}, are two fold: (1) our new depth upsampled reconstruction method formulated as a sparse edge promoting objective optimized with accelerated gradient descent to satisfy real-time requirements and (2) a new extrinsic self-calibration method based on alignments of motion from spatio-temporal information in both modalities and without position sensor requirements. Section~\ref{Sec:experimentation} presents experimental validating results to depth upsampling and motion based registration with real recorded data from the KITTI benchmark and finally Section~\ref{Sec:conclusion} concludes our findings.

\section{Proposed Approach}
\label{Sec:proposedApproach} 

This section describes the intuition behind the proposed motion guided method for the targetless extrinsic self-calibration of a multi-modal system composed of a LiDAR and a camera pair. Self-calibration is formulated as the problem of automatically finding the 6 DOF rigid body transformation of relative pose that registers or aligns data captured on-the-fly through the multiple sensing modalities and without any specific target requirements. 
Following self-calibration, a method that operates in real-time to upsample the sparse depth resulting from the re-projection of sparse LiDAR points into the camera reference is presented along with its formulation and its corresponding optimization. 

\subsection{Motion based self-calibration}
\label{ssec:registration}

\begin{figure*}[htb] 
	\centering 
	\includegraphics[width=1.0\linewidth]{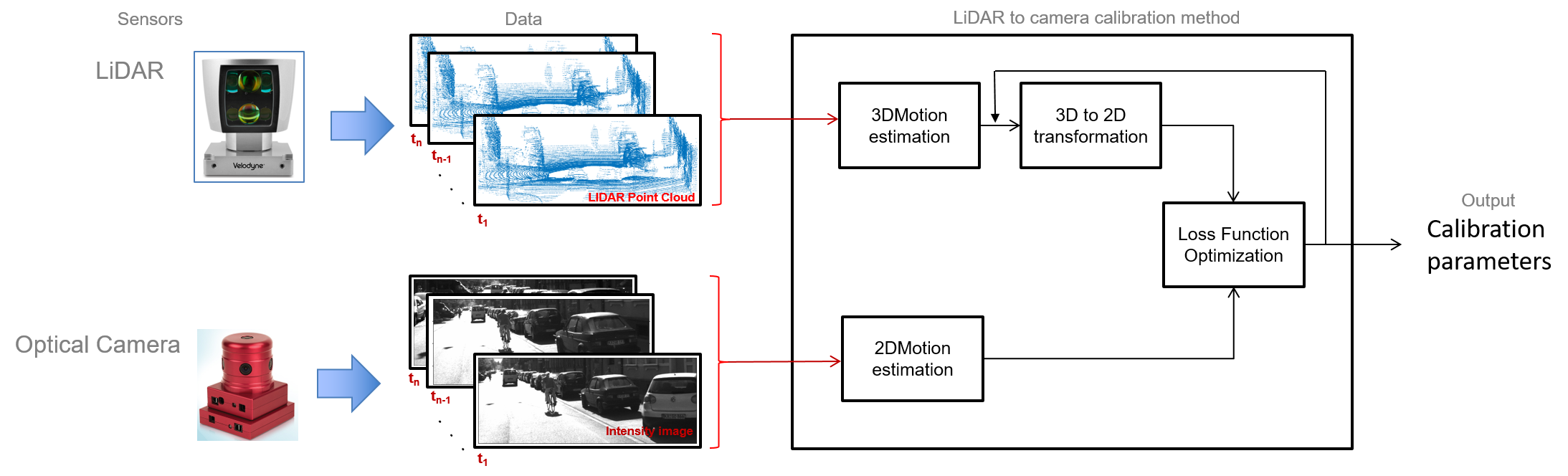}	
	\caption{Motion guided LiDAR-camera self-calibration schematic.  }
	\label{fig:schematicMotion} 
\end{figure*}
%

The presented method for the extrinsic self-calibration of a LiDAR-camera system computes 6 parameters: a roll, pitch and yaw rotations and x, y and z translations characterizing a 3D rigid body transformation. The direction in which these parameters are computed is LiDAR-to-camera while the camera-to-LiDAR direction can be computed trivially given the former. The method works with any LiDAR-camera position configuration as long as there are overlapping fields of views (FOV) between the sensors being registered.

The idea of the proposed method is simple: under alignment scene motion in the modality being registered should follow scene motion in the reference modality (e.g., camera). The advantages of registering modalities from the motion cues is two-fold: (i) it avoids the challenges related to modalities measuring different units (e.g., intensity, depth, etc) and (ii) it constraints the alignments both spatially and temporally. The procedure describing this idea and summarized in Figure \ref{fig:schematicMotion} uses a sequence of independent $L+1$ time-synchronized 2D image and 3D point cloud pairs from camera and LiDAR correspondingly as input. These sequences are then split into sequential pairs to obtain a total of $L$ pairs of 2D and 3D motion vectors computed independently for each modality. Given these, we optimize the self-calibration objective function formulated as shown in Eq. \eqref{motion}: 

\begin{equation} \label{motion}
\test = 
\argmin_{ 
	\tb \in \SE3 } 
\underbrace{
	\left \{
	\frac{1}{L} \sum\limits_{l=1}^L \left \| w(\xbi) -w(\xbd,\tb) \right \|_{\ell_2} 
	\right \}. 
}_{C(\tb)}
\end{equation}

Here, $w \in \R^{N \times 2}$ corresponds to the motion vector in the camera image plane defined by $w(x, \cdot) = [u(x, \cdot), v(x, \cdot)]^{\text{T}}$ where $u \in \R^{N}$ and $v\in \R^{N}$ are the horizontal and vertical motion operators, respectively. The term $\xbi \in \R^{N \times 2}$ represents the sequential image pair $[ \xb^l, \xb^{l+1} ]$ where $\xb^l \in \R^N$ is the column-wise vectorization of an image at the index $l$ describing a time instant. Similarly, the term $\xbd \in \R^{M \times 3 \times 2} $ represents the corresponding sequential 3D pair of point clouds composed of $[ \Xb^l, \Xb^{l+1} ]$ where $\Xb^l \in \R^{N \times 3}$ is the point cloud at time instant $l$. In words, Equation \eqref{motion} finds the optimal parameters that minimizes the distance (in the $\ell_2$ sense) between a sequence of $L$ motion vectors computed from the camera and LiDAR modalities, independently. One thing to note here is that the 3D motion vectors are re-projected into the image plane using a state of the 6DOF parameters $ \tb \in \SE3$ and the camera intrinsics and thus Eq. \eqref{motion} computes motion distances in the image plane. On another side note we would like to mention that motion distance comparisons are only performed at pixel locations occupied by valid re-projection of LiDAR measurements and that each motion vector, regardless of modality, is normalized to be of unit $\ell_2$-norm.

\begin{figure}[h] 
	\centering 
	\subfloat[Side-view ]{\includegraphics[width=0.35\linewidth]{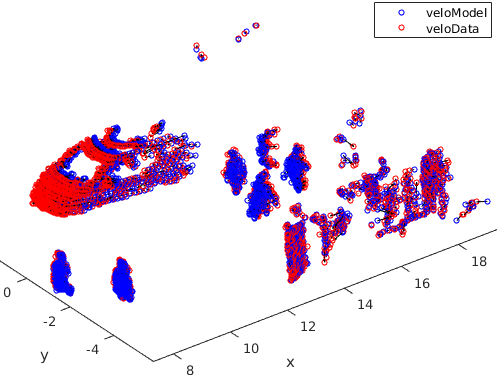}}
	\hspace{1em}
	\subfloat[Top down view]{\includegraphics[width=0.25\linewidth]{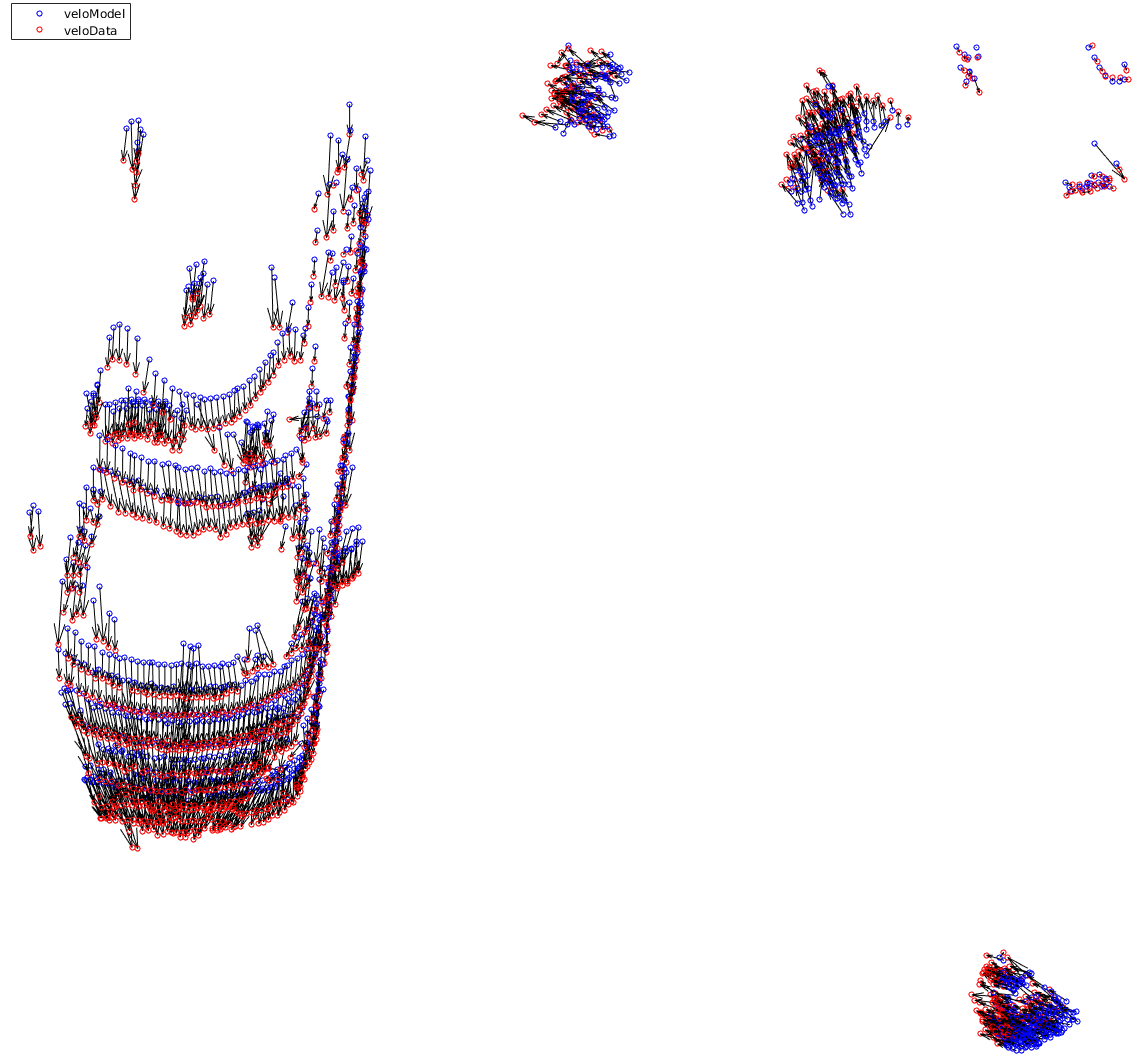}}
	%
	\hspace{1em}
	\subfloat[Pedestrian \newline motion zoomed]{\includegraphics[width=0.15\linewidth]{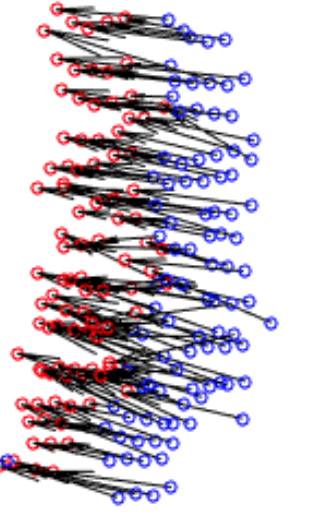}}
	\hspace{1em}
	\subfloat[Vehicle \newline motion zoomed]{\includegraphics[width=0.15\linewidth]{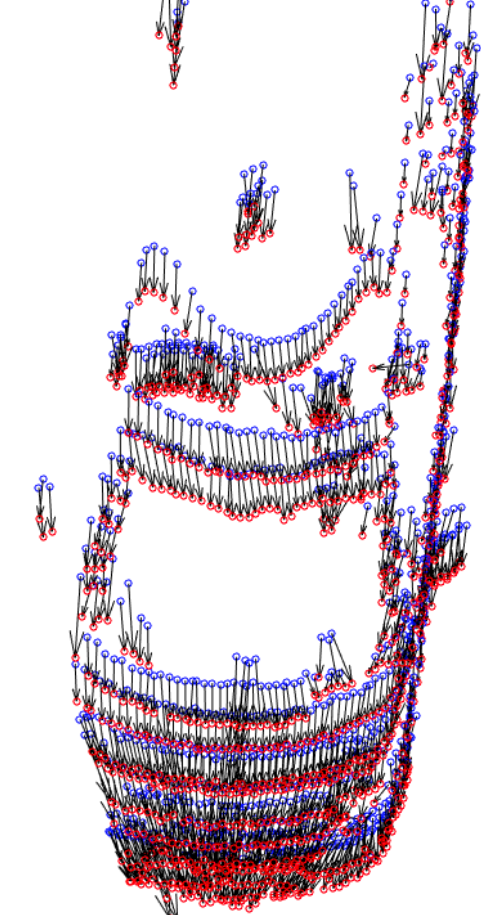}}	
	\caption{Illustrative example of the solution of 3D motion flow given a reference and target point cloud.}
	\label{fig:association} 
\end{figure}
Motion for the image sequence can be estimated here by any state of the art learning based optical flow method suitable to images such as Dosovitskiy et al. \cite{Dosovitskiy2015}. However, given that current learnable optical flow methods are not invariant to camera model and type, which is highly inconvenient to our camera-LiDAR registration problem, we resort to the total variation (TV) optical flow method of Zach \cite{Zach2007}. TV estimates motion by imposing piece-wise constant motion field constraints and has the benefit that is camera type and model invariant. For the point cloud sequence, 3D motion is estimated by sequentially taking pairs of consecutive point-clouds and computing motion in each pair along the sequence. The problem of estimating 3D motion in pairs of point clouds is posed here as a vector difference between points in the point cloud at time instant $l$ with their corresponding point associations in the point cloud at time instant $l+1$. To compute point associations between point clouds a graph-based method that optimizes for an injective mapping between points is used. Such an assignment is formulated as the optimal transport problem that minimizes the sum of distances of point associations described in Eq. \eqref{assignment}.
\begin{eqnarray} \label{assignment}
\hat{\A} = \argmin_{ \A \in \Acal } 
\left \{ \sum_{i = 1}^{N} \sum_{j = 1}^{M} \A_{i,j} \Gd_{i,j} \right \} \\ \nonumber
s.t. \quad \sum_{j = 1}^{N} \A_{i,j}  = 1 \quad \forall i \\ \nonumber
\quad \sum_{i = 1}^{M} \A_{i,j}  \leq 1 \quad \forall j \nonumber
\end{eqnarray}
Eq. \eqref{assignment} optimizes an association of points $\xb_i \in \Xb^{l} $ to points $\xb_j \in \Xb^{l+1}$ based on the cost of the Euclidean distance assignment and injective mapping constraints. The association is described by means of the matrix elements $\A_{i,j} \in \{0,1\} $ that associates points indexed by $i$ to $j$ via a non-zero entry. The matrix $\Gd \in {\R^+}^{N \times M}$ is the Euclidean distance matrix (EDM) consisting of elements computed as in Eq. \eqref{Euclidean} as:  
\begin{equation} \label{Euclidean}
	\Gd_{i,j} = \| \xb_i - \xb_j \|_{\ell_2}^2
\end{equation}
i.e., the pair-wise distance between points $\xb_i \in \Xb^{l} $ and $\xb_j \in \Xb^{l+1}$. The first constraint in Eq. \eqref{assignment} forces one point in $\Xb^{l+1} $ to be associated to only one point in $\Xb^{l}$ while the second constraint forces a point in $\Xb^{l}$ to be assigned to at most a point in $\Xb^{l+1} $, thus constraining the solution to be an injective map.
There are multiple methods to solve the problem in Eq. \eqref{assignment} suitable to the specific variation in its constraints. Here, we choose the iterative algorithmic solution of Bertsekas \cite{Bertsekas88}. A representative example illustrating the result of application of this algorithm to a pair of two consecutive points clouds is shown in Figure \ref{fig:association} only for points that underwent motion.
Here, blue points represent the point cloud $\Xb^{l} $, red points represent point cloud $\Xb^{l+1} $ and black arrows represent the motion vectors between them. Note here that no point in $\Xb^{l}$  is assigned to more than one point in $\Xb^{l+1}$. In general, we find that 3D motion computed in such way is effective and suitable to at least our registration problem and thus we use it throughout this paper to compute 3D motion.

To minimize the cost function in Eq. \eqref{motion}, we use the simplex optimization method from Nelder-Mead described in Lagarias et al. \cite{Lagarias95}. Other optimization methods could be used instead, however, it was decided to use this method because it is differentiation-free. Also, note that the chosen method is locally convergent and thus requires a rough initialization restricting search to a small neighborhood.

\subsection{Depth Upsampling}
\label{ssec:depth}

\begin{figure*} 
	\centering 
	\includegraphics[width=1.00\linewidth]{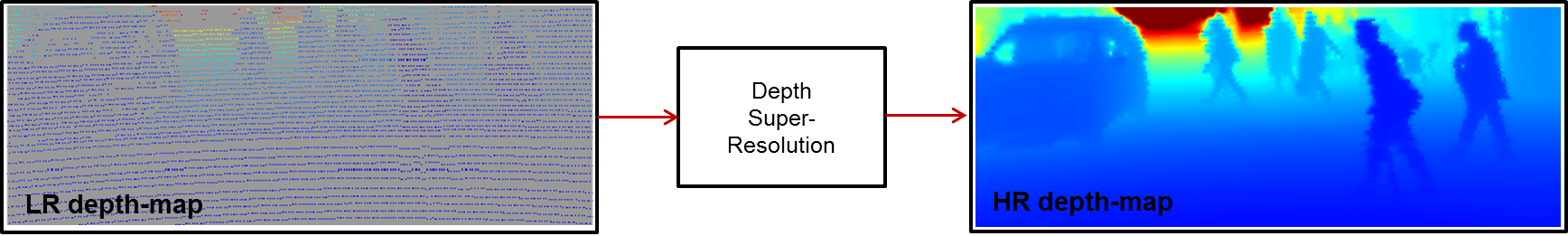}	
	\caption{Accelerated depth super-resolution reconstruction schematic.  }
	\label{fig:schematicDepthSR} 
\end{figure*}
%

The proposed method for depth upsampled reconstructions assumes availability of the extrinsic calibration parameter estimates obtained as described in Section \ref{ssec:registration}. Reprojection of the LiDAR measurements into the image plane using the found calibration parameters and the camera intrinsics results in a sparse depth-map with defined values spread non-uniformly over the image plane. The tackled problem resolves the question of how do we upsample or fill in the missing values in a sparse depth map quickly enough to make it feasible for real-time applications. In other words, how does it go from the sparse to the upsampled depth map shown in the left and right side shown in Figure \ref{fig:schematicDepthSR}, respectively. Note that the left sparse and right upsampled depth map in Figure \ref{fig:schematicDepthSR} represents missing information with gray; and the remaining color spectrum as the actual depth. 

The problem is formulated as the optimization of an objective function that minimizes an anisotropic $\ell_1$-norm on gradients subject to equality constraints with the sparsely distributed pixels containing depth values. Mathematically this can be defined as shown in Eq. \eqref{depthSR}:
\begin{equation} \label{depthSR}
\xest = 
\argmin_{ \xb \in \mathcal{X}_{\Omega} } 
\underbrace{ \sum_{k=\{x,y\}} \| \g_k \xb \|_{\ell_1} }_{f(\xb)},
\text{  s.t.   }
\xb_{\dO} = \y_{\dO}  
\end{equation}
Here, $\dO$ denotes the subset of pixels in the sparse depth-map $\y \in \R^N$ with a depth measurement from the LiDAR re-projection and $\xb$ is the upsampled depth. In Eq. \eqref{depthSR}, the gradient $\g: \R^N \rightarrow \R^{N \times 2}$ is the first order forward difference defined point-wise as shown in the Eq. \eqref{Gradient}:
\begin{equation} \label{Gradient}
[\g \xb ]_n = 
\begin{pmatrix} 
[\g_x \xb]_n \\ 
[\g_y \xb]_n 
\end{pmatrix} =
\begin{pmatrix}
[\xb]_{n + N_y} - [\xb]_n \\
[\xb]_{n+1} - [\xb]_n
\end{pmatrix}
\end{equation} 
were $ \g_x $ and $ \g_y $ denotes the horizontal and vertical components, respectively. Note that the input definitions of Eq. \eqref{depthSR} and \eqref{Gradient} assume a column-wise vectorization of a 2D image of size $ N_y \times N_x $. The interpretation of Eq. \eqref{depthSR} is as follows: $f(\xb)$ promotes depth reconstructions that minimize the strength and occurrence of depth discontinuities (i.e., of sparse depth gradient) as imposed by the $\ell_1$ norm denoted as $\|\cdot\|_{\ell_1}$ while the equality constraint ensures consistency in pixels with available depth values. Current commercially available LiDARs can offer range noise RMS at 2 cm thus supporting the high reliance of Eq. \eqref{depthSR} on the re-projected measurements. 

Since the cost function in Eq. \eqref{depthSR} is convex and non-smooth, the accelerated gradient projection method of Nesterov \cite{Nesterov1983} along with a proximal method for the non-smooth $\ell_1$-norm is used. This proximal method is the non-linear soft-thresholding operator $\eta_{\tau}(\cdot): \R^N \rightarrow \R^N $ with threshold $\tau$ defined element-wise as shown in Eq. \eqref{soft}:
\begin{equation} \label{soft}
[\eta_{\tau}(\xb) ]_n = \mbox{sgn} ( [\xb]_n ) (|[\xb]_n | - \tau )_+.
\end{equation}
Such an operator promotes sparsity of the solution discontinuities at each iteration by collapsing values near zero to zero. With such an accelerated sparse promoting optimizer, the depth reconstruction achieves a rate of convergence of $O(1/k^2)$ (where $k$ is the iteration number) due to the acceleration promoted by Nesterov's projection method. A summary of the algorithm is included in Algorithm \ref{algorithm1} 
\begin{algorithm}[t]
	\caption{ Depth upsampling reconstruction } \label{algorithm1}
	\begin{algorithmic}[1]
		\BState \textbf{input:} Sparse depth $ \y$ and learning rate $\gamma > 0$.
		\BState \textbf{set: }  iteration index $ t \leftarrow 1$, momentum variable $q_0 \leftarrow 1 $, initialization $\s^0 \leftarrow \y$ and index set of non-missing depth pixels $\Omega' = \text{supp}(y)$	
		\BState \textbf{repeat}
		\State \quad $\zO^t \leftarrow \eta_{\tau}(\sO^{t-1} - \gamma \IO \nabla f( \s^{t-1} ) ) $
		\State \quad $ q_t \leftarrow \frac{1}{2} \left ( 1 + \sqrt{1+4 q^2_{t-1}} \right )$ 
		\State \quad $ \lambda_t \leftarrow (q_{t-1} - 1)/q_t$ 		
		\State \quad $\sO^t \leftarrow (1+ \lambda_t)\zO^t - \lambda_t \zO^{t-1} $.
		\State \quad $ t \leftarrow t + 1$
		\BState \textbf{until: } stopping criterion
		\BState \textbf{set:  } $ \xest_{\Omega'} = \y_{\Omega'} $, $ \xest_{\Omega} = \sO^t $
		\State \textbf{return: } Depth reconstruction $ \xest $
	\end{algorithmic}
\end{algorithm}
when, $\IO: \R^N \rightarrow \R^M$ is the rectangular matrix populated with 1's at entries indexed by $\Omega$ and zero otherwise. This matrix is thus a selector of pixels with missing depth values. In words, in every iteration of Algorithm \ref{algorithm1} the depth values sparsely distributed along pixels propagate spatially to every pixel in the upsampled depth-map reconstruction provided this propagation satisfies the constraints imposed by the objective function. The way in which depth values are propagated is dictated by a minimization of the gradient of the objective function denoted by $\G f(x)$ and with a strength indicated by the learning parameter $\gamma$ at each iteration. Such gradient based optimization and convex nature of the cost function guarantee that at each iteration it gets closer to convergence. The additional steps in lines 5,6 and 7 of algorithm \ref{algorithm1} are Nesterov projections that further propagate depth values a little more with strength $\lambda_t$.

\section{Experimentation} 
\label{Sec:experimentation}

To validate the approaches, the KITTI benchmark
\cite{Geiger13} dataset is used. This data was collected with a vehicle outfitted with several perception and inertial sensors. The experiments use only the time synchronized data from the Velodyne HDL-64E 3D-LiDAR scanner and the left PointGray Flea2 grayscale
camera. However, the approach we propose here for both motion guided registration and depth up-sampling can be scaled to multiple cameras/LiDARs configurations. The KITTI dataset provides the extrinsic calibration parameters for each sensor mounted on the vehicle. The parameters for the LiDAR and camera pair chosen for the experiments are also given with the dataset.

\subsection{Motion guided self-calibration}
\begin{figure*} [!t]
	\centering 
	\subfloat[Roll offset]{\includegraphics[width=0.5\linewidth]{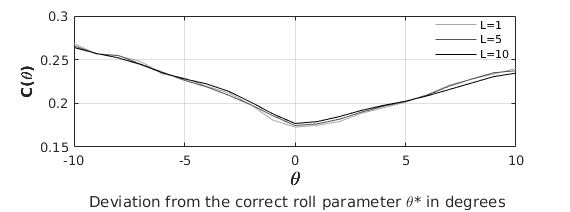}}
	\subfloat[Pitch offset]{\includegraphics[width=0.5\linewidth]{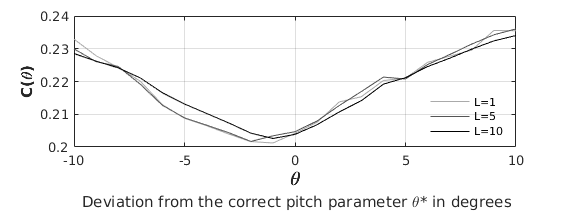}}
	
	\subfloat[Yaw offset]{\includegraphics[width=0.5\linewidth]{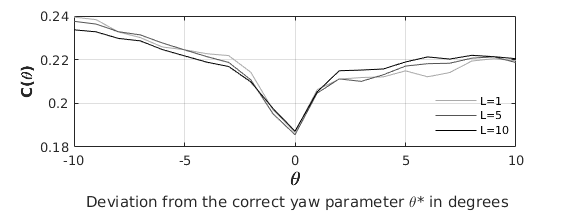}}
	\subfloat[X offset]{\includegraphics[width=0.5\linewidth]{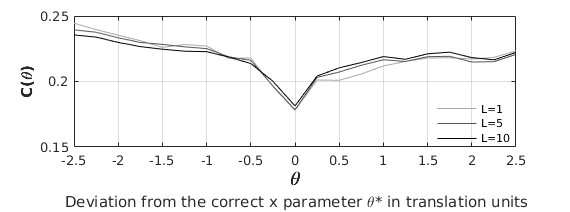}}
	
	\subfloat[Y offset]{\includegraphics[width=0.5\linewidth]{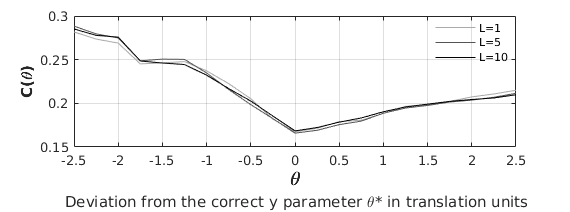}}
	\subfloat[Z offset]{\includegraphics[width=0.5\linewidth]{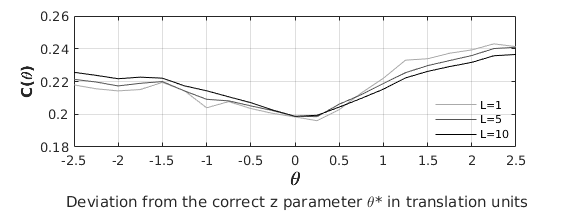}}
	\caption{Effect of parameter offset in Eq. \eqref{motion}. a) Roll,
		b) Pitch, c) Yaw, d) X, e) Y, f) Z.}
	\label{fig:motionCals} 
\end{figure*}
The first experiment presented here on motion guided self-calibration shows the behavior of the objective function in Section \ref{ssec:registration} as a function of an offset from the true extrinsic parameters provided by KITTI. The data from KITTI used in this regard corresponds to a vehicle driving in an urban area. We manually choose corresponding image (converted to grey scale) and point cloud sequences of 2,6 and 11 consecutive frames that contain motion in the scene and evaluate the objective function in Section \ref{ssec:registration} by fixing 5 parameters at true value and offset one to different values and then repeat the process for all 6 DOF of roll, pitch, yaw and x, y, z. A representative example is shown in Figure \ref{fig:motionCals}, which includes plots of the behavior of Eq. \eqref{motion} for a given parameter offset.
It was observed that the proposed objective function shows a clear global minimum at the correct calibration parameters with a few minor local bumps along the way. These bumps are further smoothed as the number of frames $L$ used in Eq. \eqref{motion} is increased pushing our cost function to be both convex and smooth in the local neighborhood around the optimal value.

\subsubsection{Convergence of the cost function}
\begin{figure} 
	\centering 
	\subfloat[L=5 Motion vector pairs]{\includegraphics[width=0.5\linewidth]{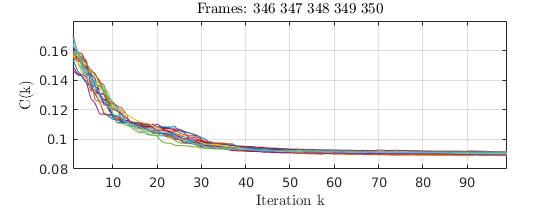}}
	\caption{Convergence as a function of iterations.}
	\label{fig:simplexConvergence} 
\end{figure}

This experiment illustrates the convergence behavior of the simplex optimization of Eq. \eqref{motion}. For this purpose,  the simplex optimization approach was initialized with 7 points where these were chosen randomly within
$20\%$ offset from the true. A representative example of the convergence of the optimization is illustrated in Figure \ref{fig:simplexConvergence}, showing the first 100 iterations of 20 independent experiments for the $L=5$ case. This convergence behavior is consistent throughout different scenes in an urban environment, driving behavior and number of motion vectors $L$. However, one of the limiting cases for convergence of the algorithm is when $L \rightarrow 1$ where along with cases of no vehicle nor scene dynamics results in motion vectors with no information to constrain data alignments. A side from these, convergence presented a rate behavior similar to that in \ref{fig:simplexConvergence} with prominent damping at 50 iterations.

\subsubsection{Robustness of registration algorithm}
This experiment shows the performance and robustness of the proposed motion based registration formulation against randomized initializations and compare this to the best possible registration from robust perspective n-point (RPnP) algorithm of Li et al. \cite{Li2012} when there is a human doing manual correspondences. A total of 100 experiments were conducted where in each registration using either the proposed method or RPnP a sequence of consecutive and time-synchronized 6 gray scale image/point cloud pairs where used (i.e., $L=5$ for the proposed method). In each of the 100 independent registration trials, different scenes from an urban scenario with significant pedestrian traffic were randomly selected. The registration parameters of these trials were randomly initialized within offsets of $ \pm 20^{\text{o}} $ for rotations and $\pm3.5$ m for translations from ground truth. Here, ground truth corresponds to the extrinsics provided from the KITTI benchmark. 
\begin{figure*}[t] 
	\centering 
	\subfloat[Rotation]{\includegraphics[width=0.5\linewidth]{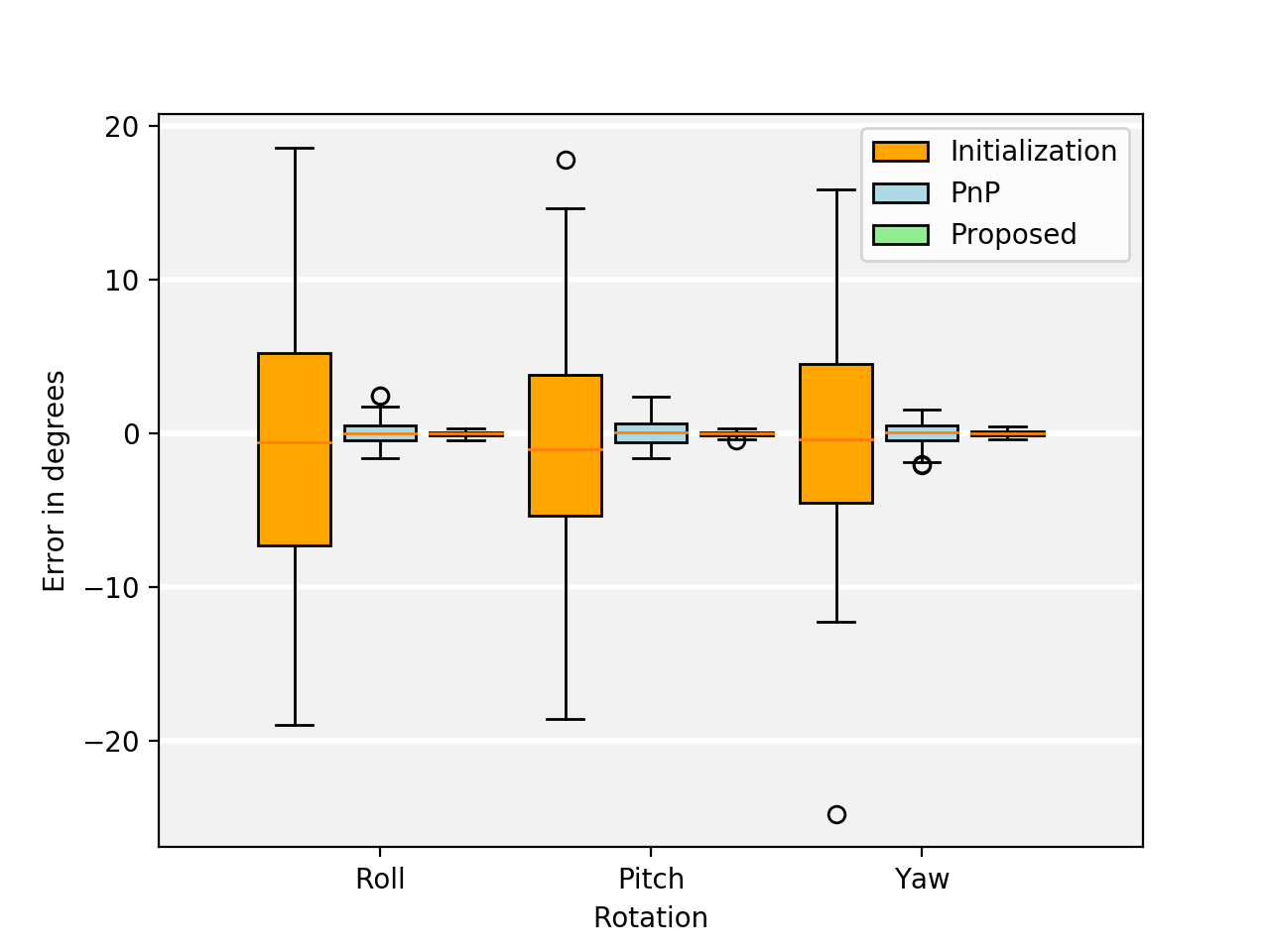}}
	\subfloat[Translation]{\includegraphics[width=0.5\linewidth]{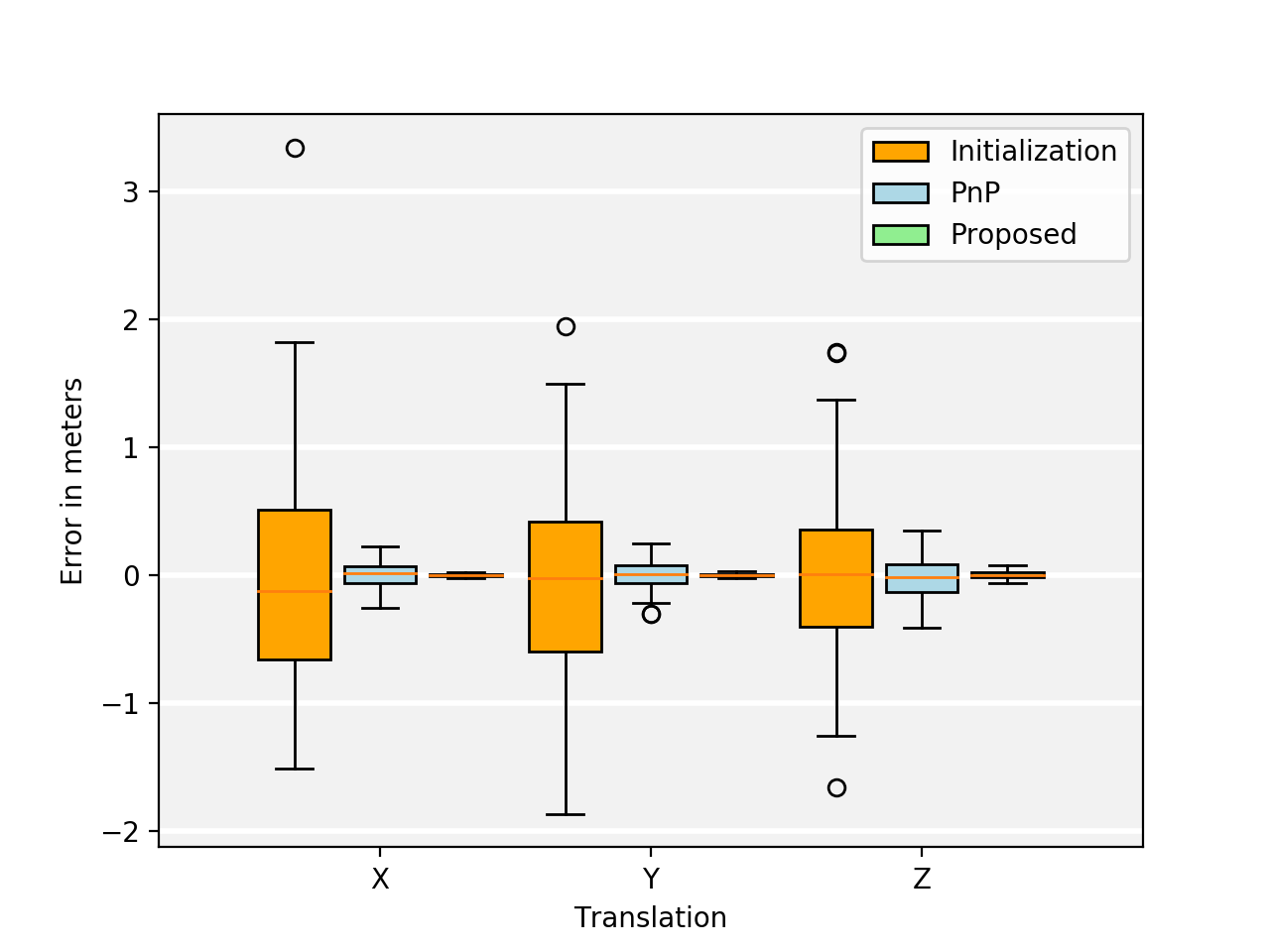}}
	\caption{Robustness of the motion guided
		registration algorithm. a) Rotation error results, b) Translation error results. Box-plots corresponding to the "Initialization" label is the random initialization distribution centered at ground truth, label corresponding to "PnP" is the calibration result of the RPnP method \cite{Li2012} while the label "Proposed" is the error performance of the proposed calibration method. }
	\label{fig:performanceRegistration}
\end{figure*}

For registration using the RPnP algorithm, a specialized user was presented with 6 image/point clouds pairs and 4 corresponding keypoints in each pair that were clearly visible were manually selected in both modalities (e.g., corners). After selection of keypoints, the user validated the selected 24 points per sequence, ran the RPnP algorithm and computed the root mean squared error (RMSE) of the result compared to the ground truth. The proposed method used in each of the 100 trials the same sequence of 6 image/point clouds pairs used in the corresponding trial for the RPnP. 

\begin{table} [h]
	\caption{\label{tab:table1} RMSE comparisons.}
	\centering
	\begin{tabular}{lcccccc}
		\hline
		& & & & & &\\
		& \multicolumn{3}{c}{Rotation ($^{\circ}$)} &  \multicolumn{3}{c}{Translation (m)}   \\
		\multirow{2}{*}{Method} & \multirow{2}{*}{Roll} &  \multirow{2}{*}{Pitch}  & \multirow{2}{*}{Yaw} & \multirow{2}{*}{X} &  \multirow{2}{*}{Y} & \multirow{2}{*}{Z}\\
		& & & & & &\\
		\hline
		\\
		Initialization & 8.39 & 7.37 & 7.02 & 0.86 & 0.80 & 0.68\\
		PnP \cite{Li2012} & 0.74 & 0.88 & 0.86 & 0.10 & 0.11 & 0.15 \\
		Proposed & \pmb{0.15} & \pmb{0.15} & \pmb{0.17} & \pmb{0.01} & \pmb{0.01} & \pmb{0.03}\\
		\hline
	\end{tabular}
\end{table}

Figure \ref{fig:performanceRegistration} summarize the performance results obtained from the experimentation with corresponding RMSE's listed in Table \ref{tab:table1}. The whiskers in the box-plots of Figure \ref{fig:performanceRegistration} corresponds to the range of the error while the boxplots shown in a color extend from the 1st to 3rd quartile of the error distribution. A total of three box-plots are shown per each of the six registration parameters corresponding to the random initialization labeled as "initialization", results of RPnP \cite{Li2012} algorithm labeled as "PnP' and the proposed method labeled as "Proposed". For these experiments a total of 100 trials where run for each of the registration parameters independently. Note here that in each trial not only the parameter plotted was randomized but all 6 DOF and that error is measured here as the $\ell_2$-norm between result and ground truth. From Figure \ref{fig:performanceRegistration} and Table \ref{tab:table1} we note that both algorithms were able to reduce the random initialization error to close to zero across all trials and for all 6 parameters. Both RPnP and the proposed algorithm achieved an overall rotation RMSE's of $< 1^{\text{o}} $ and $<15$ cm for the translation. However, the proposed registration formulation outperforms the RPnP method by reducing error to within RMSE accuracies of $<0.2^{\text{o}}$ of rotation and $<3$ cm translation.

\subsection{Depth super-resolution: qualitative analysis.}

\begin{figure}
	
	\subfloat[Image from camera]{\includegraphics[width=0.5\linewidth]{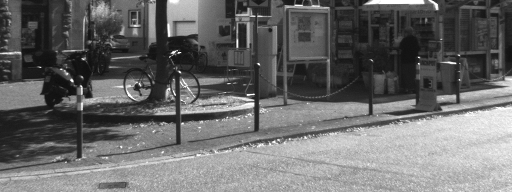}}
	\subfloat[Sparse reprojected LiDAR]{\includegraphics[width=0.5\linewidth]{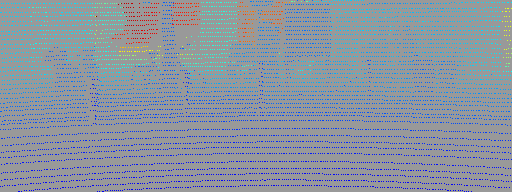}}
	
	\subfloat[TGV from Ferst \cite{Ferstl13}]{\includegraphics[width=0.5\linewidth]{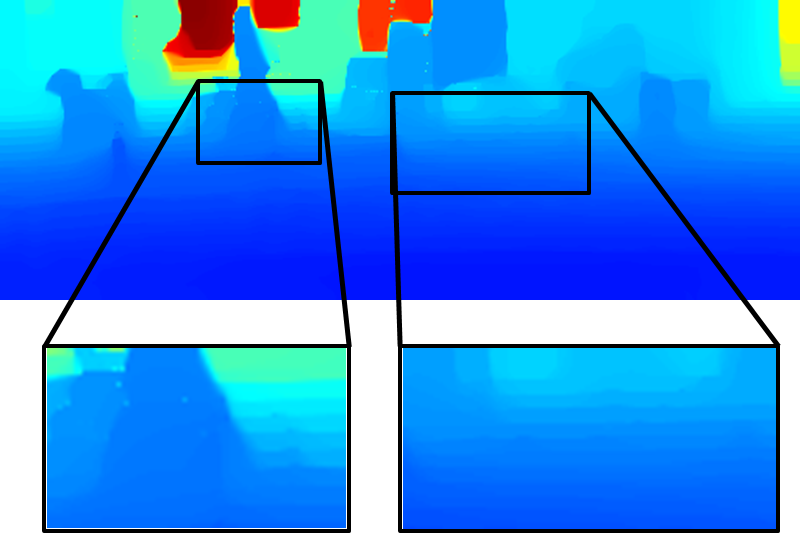}}
	\subfloat[Proposed up-depth sample]{\includegraphics[width=0.5\linewidth]{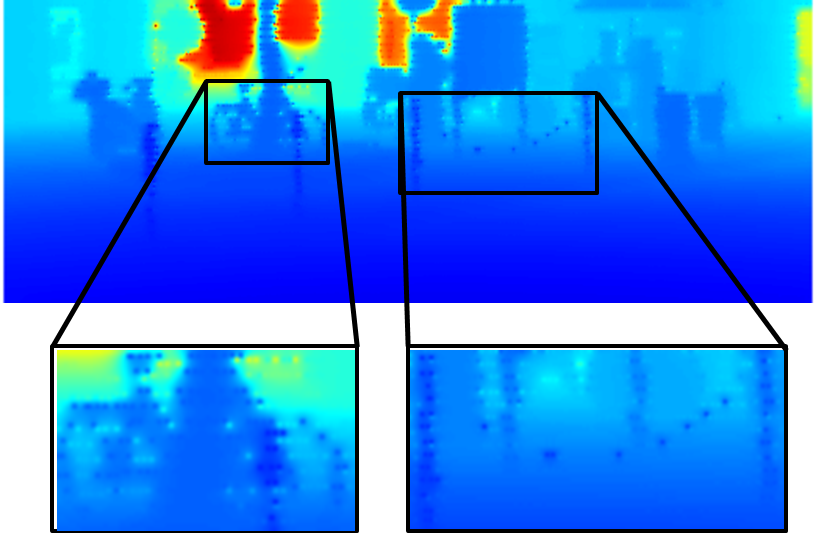}}
	
	\caption{Upsampled depth reconstruction comparison. Note in the zoomed patches in (c) and (d) showing a bike and a chain between two poles that our approach is capable of resolving finer details compared to TGV. }
	\label{fig:depthReconstructionDetails}
	
\end{figure}
\begin{figure*} [htb]
	\centering 
	\subfloat[Image from camera]{\includegraphics[width=0.25\linewidth]{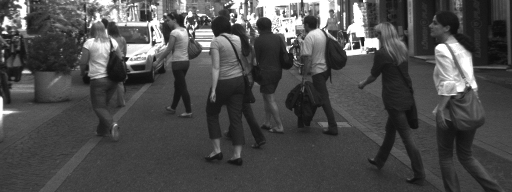}}
	\subfloat[Sparse reprojected LiDAR]{\includegraphics[width=0.252\linewidth]{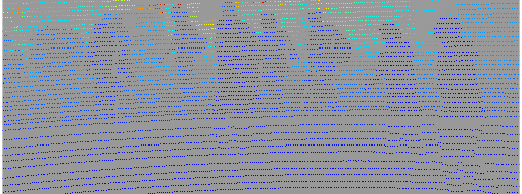}}
	\subfloat[TGV from Ferst \cite{Ferstl13}]{\includegraphics[width=0.25\linewidth]{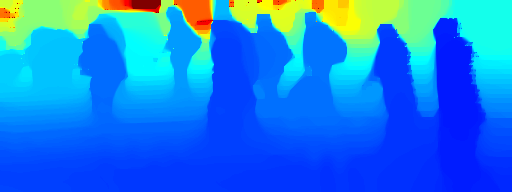}}
	\subfloat[Proposed method]{\includegraphics[width=0.25\linewidth]{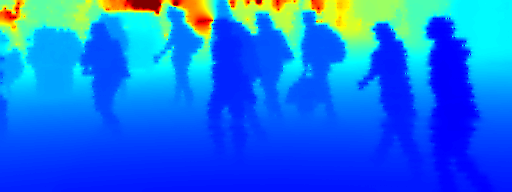}}
	\vspace{-1em}
	
	\subfloat[Image from camera]{\includegraphics[width=0.25\linewidth]{Slice82Img}}
	\subfloat[Sparse reprojected LiDAR]{\includegraphics[width=0.25\linewidth]{Slice82Depth}}
	\subfloat[TGV from Ferst \cite{Ferstl13}]{\includegraphics[width=0.252\linewidth]{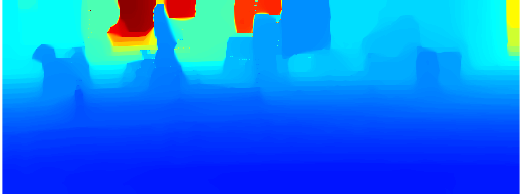}}
	\subfloat[Proposed method]{\includegraphics[width=0.252\linewidth]{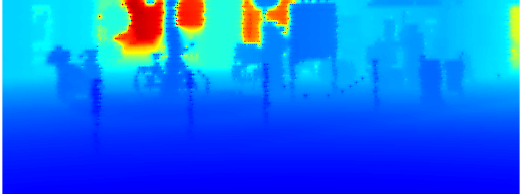}}
	\vspace{-1em}
	
	\subfloat[Image from camera]{\includegraphics[width=0.25\linewidth]{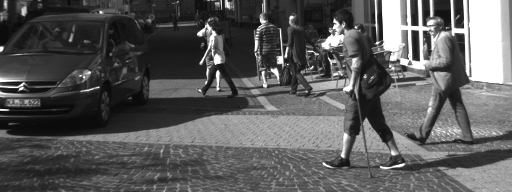}}
	\subfloat[Sparse reprojected LiDAR]{\includegraphics[width=0.253\linewidth]{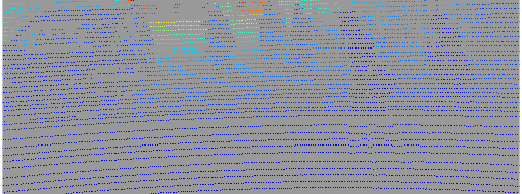}}
	\subfloat[TGV from Ferst \cite{Ferstl13}]{\includegraphics[width=0.25\linewidth]{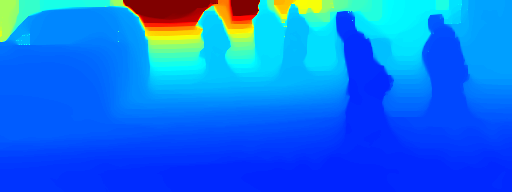}}
	\subfloat[Proposed method]{\includegraphics[width=0.25\linewidth]{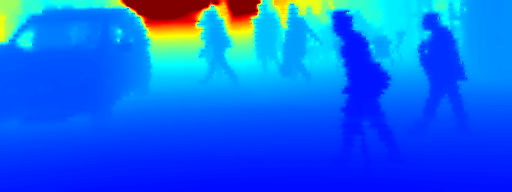}}
	\vspace{-1em}
	
	\subfloat[Image from camera]{\includegraphics[width=0.25\linewidth]{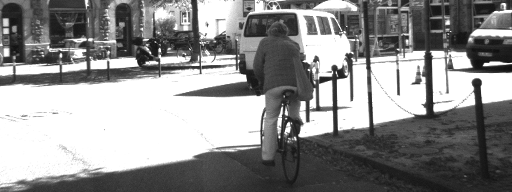}}
	\subfloat[Sparse reprojected LiDAR]{\includegraphics[width=0.253\linewidth]{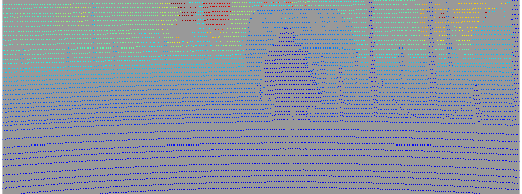}}
	\subfloat[TGV from Ferst \cite{Ferstl13}]{\includegraphics[width=0.25\linewidth]{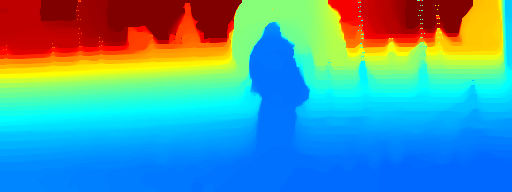}}
	\subfloat[Proposed method]{\includegraphics[width=0.25\linewidth]{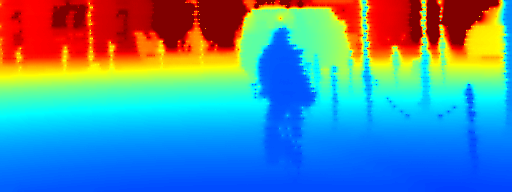}}
	\vspace{-1em}
	
	\subfloat[Image from camera]{\includegraphics[width=0.25\linewidth]{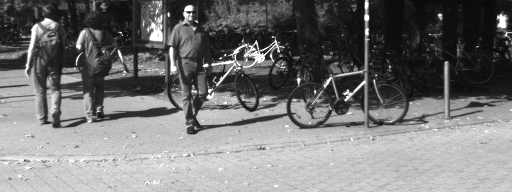}}
	\subfloat[Sparse reprojected LiDAR]{\includegraphics[width=0.252\linewidth]{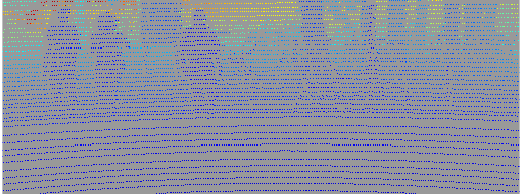}}
	\subfloat[TGV from Ferst \cite{Ferstl13}]{\includegraphics[width=0.25\linewidth]{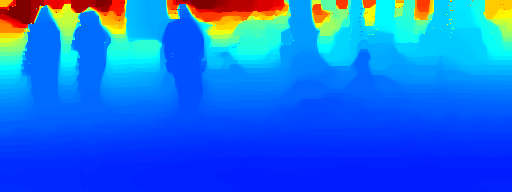}}
	\subfloat[Proposed method]{\includegraphics[width=0.25\linewidth]{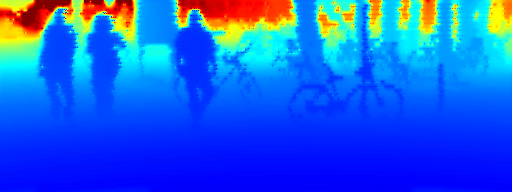}}
	\vspace{-1em}
	
	\subfloat[Image from camera]{\includegraphics[width=0.25\linewidth]{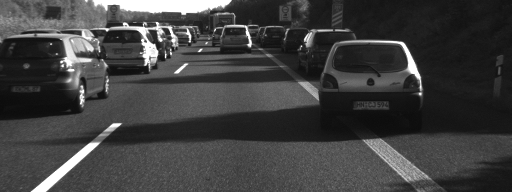}}
	\subfloat[Sparse reprojected LiDAR]{\includegraphics[width=0.252\linewidth]{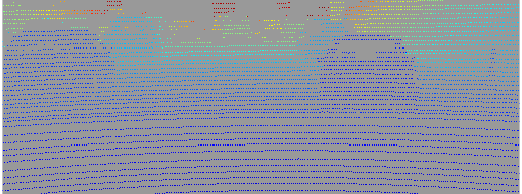}}
	\subfloat[TGV from Ferst \cite{Ferstl13}]{\includegraphics[width=0.252\linewidth]{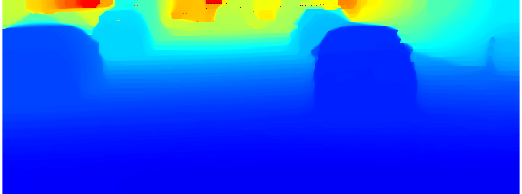}}
	\subfloat[Proposed method]{\includegraphics[width=0.252\linewidth]{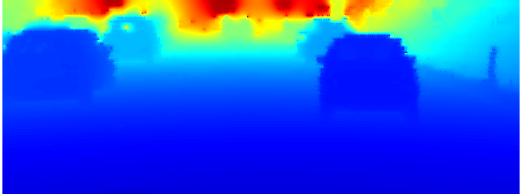}}

	\caption{Up-sampled depth reconstruction comparisons. Note in the pedestrian extremities of (d,l), posts in (h,p), bikes and other objects in (h,t) that our approach is capable of resolving finer details compared to TGV.}
	
	\label{fig:depthReconstructions} 
\end{figure*}

This experiment presents a qualitative analysis of the proposed depth upsampled
reconstruction algorithm and compare it with the generalized total variation (TGV) approach described in Ferst et al. \cite{Ferstl13}. Throughout the experiments, the proposed reconstruction uses algorithm \ref{algorithm1} with a learning rate $\gamma = 0.1$. Note that other $\gamma$ values also work well. However, that value was chosen since it was found experimentally that it gives a good trade-off between depth-map reconstruction quality and convergence time. 

To illustrate the performance comparison, Figure \ref{fig:depthReconstructionDetails} shows the capabilities of our method to resolve finer details by using only the sparse depth from LiDAR in Figure \ref{fig:depthReconstructionDetails}.b as opposed to TGV that in addition uses the image in \ref{fig:depthReconstructionDetails}.a from the camera. In Figure \ref{fig:depthReconstructionDetails}.b the gray colored pixels represent pixels with missing depth measurements from the re-projection of LiDAR data while the remaining colors represent actual depth. Comparing Figures \ref{fig:depthReconstructionDetails}.c and d show that the new method is able to resolve finer details as further illustrated in the zoomed patches showing the bike and its wheels in the bottom left patch and the chain in between poles as seen in the bottom right patch whereas these are hardly distinguishable in the TGV method.

Figure \ref{fig:depthReconstructions} also illustrates six additional depth upsampled reconstruction examples each presented in a row. First column of Figure \ref{fig:depthReconstructions} represents an image of the scene captured from the camera and converted into gray scale. The second column of Figure \ref{fig:depthReconstructions} represents the sparse depth map while the third and fourth columns the upsampled reconstructions from TGV and the proposed algorithm, respectively. Note that the proposed approach results in better reconstructions and avoids oversmoothing edges in comparison to TGV. This is specially notorius in objects with sharp edge discontinuities as for example the pedestrian legs in Figures \ref{fig:depthReconstructions}(c) versus (d) and in (k) versus (l), in the poles in (g) versus (h) and in (o) versus (p). Note also in (s) that details in trees, poles and bikes specially those in the right part of the scene are hard to resolve whereas in (t) one is able to see the tree trunks, poles and bikes present in the scene. In addition to the edge sharpness gain, the depth upsampled reconstruction by the proposed implementation took 0.033 secs per frame versus the $\sim4.4$ of the TGV method that makes it suitable for real-time applications such as AV's. This advantage is due to both Nesterov's optimization acceleration and the fixed equality constraint in Eq. \eqref{depthSR} as opposed to its corresponding relaxation in the TGV method. Finally, it is worth noting that some points that appear as artifacts at the edges in this depth reconstruction are not caused by the proposed depth upsampling method but rather by the LiDAR scanning mechanism. In particular, this is caused from the horizontal difference of laser firing/detection timings across the lasers vertically oriented inside the LiDAR that causes each to fire to a slightly different horizontal location than its vertically oriented laser neighbors.

\section{Conclusion}
\label{Sec:conclusion}
This work proposed both a novel motion based self-calibration method for multi-modal LiDAR and camera sensors and a method to generate depth upsampled reconstructions from sparsely measured re-projected LiDAR data. The experiments on recorded real-data show that the proposed motion based registration function performs well as a similarity metric between the re-projected LiDAR and camera measurements. It is efficient in constraining alignments both spatially and temporally and effective in decoupling alignments from the modality differences between intensity and 3D measurements. An additional benefit is that the newer method requires no targets and is capable of self-calibrating on the fly. Such properties, are highly attractive to general mobile autonomous robots and self-driving cars for which current practices include calibration in specialized target-based calibration facilities often expensive and impractical for mass production. The results also indicate that the proposed method for depth upsampled reconstruction achieves state of the art performance for the real-time deployment category. The algorithm renders depth reconstructions preserving smooth regions while respecting sharp edge discontinuities. The balance between performance and computational complexity of our depth upsampling makes it an attractive method perfect for mobile robotics applications imposing real-time requirements.

\bibliographystyle{IEEEbib}

\end{document}